\documentclass[letterpaper]{article} 
\usepackage{aaai25}  
\usepackage{times}  
\usepackage{helvet}  
\usepackage{courier}  
\usepackage[hyphens]{url}  
\usepackage{natbib}  
\usepackage{caption} 
\usepackage{url}
\usepackage{graphicx} 

\usepackage{microtype}
\usepackage{subfigure}
\usepackage{algpseudocode}

\usepackage{amssymb}
\usepackage{epstopdf, color, array}
\usepackage{epsfig}  
\usepackage{transparent}
\usepackage{amsmath,bm}
\usepackage{mathrsfs} 
\usepackage[english]{babel}
\usepackage{blindtext}
\usepackage{amsfonts}%
\usepackage{amsthm}%
\usepackage{mathtools}
\usepackage{mathrsfs}%
\usepackage[title]{appendix}%
\usepackage{xcolor}%
\usepackage{textcomp}%
\usepackage{manyfoot}%
\usepackage{adjustbox}
\usepackage{listings}%

\usepackage{algorithm}

\usepackage{newfloat}
\usepackage{listings}
\DeclareCaptionStyle{ruled}{labelfont=normalfont,labelsep=colon,strut=off} 
\floatstyle{ruled}
\newfloat{listing}{tb}{lst}{}
\floatname{listing}{Listing}

\title{Generalization of Compositional Tasks with Logical Specification via Implicit Planning}
\author{
    Duo Xu,
    Faramarz Fekri    
}
\affiliations{
    School of ECE \\
    Georgia Institute of Technology
    \\

    225 North Avenue, Atlanta, GA 30332 USA\\
    dxu301@gatech.edu
}

\usepackage{bibentry}

\begin{document}

\maketitle

\begin{abstract}
In this study, we address the challenge of learning generalizable policies for compositional tasks defined by logical specifications. These tasks consist of multiple temporally extended sub-tasks. Due to the sub-task inter-dependencies and sparse reward issue in long-horizon tasks, existing reinforcement learning (RL) approaches, such as task-conditioned and goal-conditioned policies, continue to struggle with slow convergence and sub-optimal performance in generalizing to compositional tasks. To overcome these limitations, we introduce a new hierarchical RL framework that enhances the efficiency and optimality of task generalization. At the high level, we present an implicit planner specifically designed for generalizing compositional tasks. This planner selects the next sub-task and estimates the multi-step return for completing the remaining task to complete from the current state. It learns a latent transition model and performs planning in the latent space by using a graph neural network (GNN). Subsequently, the high-level planner’s selected sub-task guides the low-level agent to effectively handle long-horizon tasks, while the multi-step return encourages the low-level policy to account for future sub-task dependencies, enhancing its optimality. We conduct comprehensive experiments to demonstrate the framework’s advantages over previous methods in terms of both efficiency and optimality.\footnote{Source codes will be released upon acceptance}
\end{abstract}

\section{Introduction}
\label{sec:intro}
In real-world applications, such as robotics and control system, task completion often involves achieving multiple subgoals that are spread over time and must follow user-specified temporal order constraints. For instance, a service robot on a factory floor may need to gather components in specific sequences based on the product being assembled, all while avoiding unsafe conditions. These complex tasks are defined through logic-based compositional languages, which have long been essential for objective specification in sequential decision-making \citep{de2013linear}.

Using domain-specific properties as propositional variables, formal languages like Linear Temporal Logic (LTL) \citep{pnueli1977temporal} and SPECTRL \citep{jothimurugan2019composable} encode intricate temporal patterns by combining these variables with temporal operators and logical connectives. These languages provide clear semantics and support semantics-preserving transformations into deterministic finite-state automata (DFA), which reveal the discrete structure of an objective to a decision-making agent. Generalizing across multiple tasks is crucial for deploying autonomous agents in various real-world scenarios \citep{taylor2009transfer}. In this work, we tackle the problem of generalizing compositional tasks where, at test time, the trained agent is given a DFA description of an unseen task and is expected to accomplish the task without further training.

While reinforcement learning (RL) algorithms have achieved remarkable success across numerous fields \citep{mnih2015human,wang2022deep}, they still face challenges in generalizing to compositional tasks, which differ significantly from typical problems addressed by conventional RL methods. Previous works on compositional task generalization \citep{kuo2020encoding,araki2021logical,vaezipoor2021ltl2action,den2022reinforcement,liu2022skill} trained generalizable agents with satisfying success rate.
Some approaches \citep{araki2021logical,den2022reinforcement,liu2022skill} tackle unseen compositional tasks by leveraging trained reusable skills or options. However, these methods train each option to achieve a specific subgoal independently, often without considering the task as a whole, thereby risking the loss of global optimality in task completion. Additionally, methods that train policies directly conditioned on task formulas \citep{kuo2020encoding,vaezipoor2021ltl2action} tend to exhibit slow convergence in complex tasks or environments, as they lack task decomposition and miss out on the compositional structure of such tasks. 
Detailed discussion on related works is in Appendix \ref{sec:related}.

\begin{figure}
    \centering
    \subfigure[Myopic solution]{
        \centering
        \includegraphics[width=1.7in]{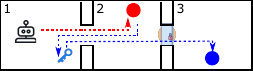}
        \label{fig:reply_fig1_1}
    }
    
    \subfigure[Optimal solution]{
        \centering
        \includegraphics[width=1.7in]{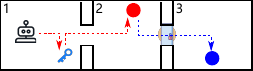}
        \label{fig:reply_fig1_2}
    }
    \caption{\small Motivating example 1. Task: first go to red ball, and then blue ball. Red: reaching red ball. Blue: reaching blue ball.}
    \label{fig:reply_fig}
\end{figure}

\noindent
{\bf Motivating Examples.} The first example, shown in Figure \ref{fig:reply_fig}, involves a robot starting in room 1 with a locked door between rooms 2 and 3. This door can only be opened using the key located in room 1. The task is to visit the red ball first, then the blue ball. Previous option-based methods train options for reaching each ball independently, disregarding dependencies between subgoals. This approach results in the myopic solution illustrated in Figure \ref{fig:reply_fig1_1}, where the robot wastes additional steps retrieving the key compared to the optimal solution shown in Figure \ref{fig:reply_fig1_2}.

The second example, illustrated in Figure \ref{fig:imp-plan-diagram} with the task automaton on the right, assigns rewards of 1 and 10 for reaching areas "a" and "c", respectively, while other areas yield no rewards. The optimal solution for completing the task with maximal rewards is to reach "b" first, then "c", while avoiding "a" and "d". However, if task-conditioned policies are trained using previous methods \citep{kuo2020encoding,vaezipoor2021ltl2action}, the agent often ends up reaching "a" and avoiding "d" since it is easier and still completes the task, producing sub-optimal solution for the given task. This occurs because the task-conditioned policy is not explicitly trained to avoid "a" while reaching "b" at the initial automaton state $0$ as this avoidance requirement is not clearly specified by the task automaton. The resulting sub-optimality arises from overlooking subgoal dependencies: reaching "a" from the state $0$ can directly accomplish the task, making "b" and "c" inaccessible and hence getting sub-optimal rewards for completing the task.

\begin{figure}
    \centering
    \fontsize{8pt}{8pt}\selectfont
    \def\svgwidth{2.35in}
    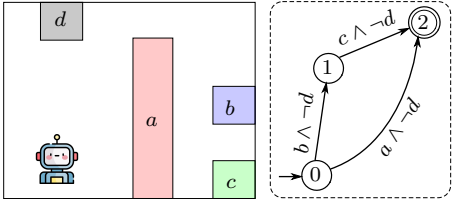
    \caption{\small Motivating example 2. Left: map. Right: task automaton with self loops omitted. }
    \label{fig:imp_mot_exp2}
\end{figure}

In this work, we introduce a hierarchical RL framework designed for zero-shot generalization, enabling a trained agent to tackle new compositional tasks specified in logic without additional training. Unlike previous methods, our approach incorporates a novel implicit planner as the high-level module. This planner selects the next sub-task and estimates the multi-step return for completing the remaining tasks, thereby guiding the low-level agent while accounting for dependencies among future sub-tasks. The low-level module is responsible for choosing primitive actions to accomplish the assigned sub-task, with its decisions conditioned on both the current and upcoming sub-tasks required to complete the task.

In experiments, we demonstrate the advantages of the proposed framework over baselines in three environments, including both discrete and continuous state and action spaces. Based on comprehensive experiments, we show the proposed framework outperforms baselines in terms of both optimality and learning efficiency.

\section{Preliminaries}
\label{sec:prel}

\subsection{Episodic Labeled MDP}
\label{sec:rl}
The proposed framework is to solve tasks in an episodic labeled Markov decision process (MDP) $\mathcal{M}$ defined as a tuple $(\mathcal{S},\mathcal{A},P,R,\mathcal{S}_0,\gamma,H,\mathcal{P}_0,L)$ \citep{sutton2018reinforcement,icarte2018using,jothimurugan2019composable} where $\mathcal{S}$ is the state space, $\mathcal{A}$ is the set of actions or action space, $P(s'|s,a)$ is the transition probability of reaching next state $s'$ from $s\in\mathcal{S}$ given action $a\in\mathcal{A}$, $R:\mathcal{S}\times\mathcal{A}\times\mathcal{S}\to\mathbb{R}$ denotes the reward function, $\mathcal{S}_0$ is the initial state distribution, $\gamma\in[0,1]$ is the discount factor, $H$ is the maximum length of each episode, $\mathcal{P}_0$ is the set of atomic propositions, and $L:\mathcal{S}\to 2^{\mathcal{P}_0}$ is the labeling function mapping any state $s\in\mathcal{S}$ to a finite set of propositions in $\mathcal{P}_0$ holding true in state $s$. 
When interacting with the environment, given the current state $s$ and task $\phi$, the agent selects action according to its policy, i.e., $\pi(\cdot|s,\phi)$, and it makes progress towards completing $\phi$ by observing symbols $L(s')$ of next state. In each episode, the agent's target is to complete the task $\phi$ within $h (\le H)$ steps and maximize the accumulated rewards ($G=\sum_{t=0}^h \gamma^t R(s_t,a_t,s_{t+1})$) at the same time.

\subsection{Task Specification Language}
\label{sec:ltl}
A compositional task considered in this work is described by a logic specification formula  $\phi$, a Boolean function that determines whether the objective formula is satisfied by the given trajectory or not \citep{pnueli1977temporal}. In this work, we adopt the specification language SPECTRL \citep{jothimurugan2019composable} to express the logic and temporal relationships of subgoals in tasks. A specification $\phi$ in SPECTRL is a logic formula applied to trajectories, determining whether a given trajectory $\zeta=(s_0,s_1,\ldots)$ successfully accomplishes the task specified by $\phi$. For rigor of math, $\phi$ can be described as a function $\phi:\mathcal{Z}\to\{0, 1\}$ producing binary outputs, where $\mathcal{Z}$ is the set of all the trajectories.

Specifically, a specification is defined based on a set of atomic propositions $\mathcal{P}_0$. For each proposition $p\in\mathcal{P}_0$, the MDP state $s$ of the agent satisfies $p$ (denoted as $s\models p$) when $p\in L(s)$ and $L$ is labeling function introduced in Section \ref{sec:rl}. The set of symbols $\mathcal{P}$ is composed by conjunctions of atomic propositions in $\mathcal{P}_0$. 

Based on definitions above, the grammar for formulating SPECTRL specifications can be written as:
\begin{equation}
    \phi::=\text{ achieve } b \hspace{5pt}|\hspace{5pt} \phi_1 \text{ ensuring } b \hspace{5pt}|\hspace{5pt} \phi_1;\phi_2 \hspace{5pt}|\hspace{5pt} \phi_1 \text{ or } \phi_2 \label{spectrl}
\end{equation}
where $b\in\mathcal{P}$. Here "achieve" and "ensuring" correspond to "eventually" and "always" operators in LTL \citep{pnueli1977temporal,araki2021logical}. Given any finite trajectory $\zeta$ with length $h$, the satisfaction of a SPECTRL specification are defined as:
\begin{enumerate}
    \item $\zeta\models \text{ achieve } b$ if $\exists i\le h, s_i\models b$ (or $b\in L(s_i)$)
    \item $\zeta\models \phi \text{ ensuring } b$ if $\zeta\models\phi$ and $\forall i\le h, s_i\models b$
    \item $\zeta\models \phi_1;\phi_2$ if $\exists i<h, \zeta_{0:i}\models \phi_1$ and $\zeta_{i+1;h}\models\phi_2$
    \item $\zeta\models\phi_1 \text{ or } \phi_2$ if $\zeta\models\phi_1$ or $\zeta\models \phi_2$
\end{enumerate}
Specifically, the statement 1) signifies that the trajectory should eventually reach a state where the symbol $b$ holds true. The statement 2) means that the trajectory should satisfy specification $\phi$ while always remaining in states where $b$ is true. The statement 3) signifies that the trajectory should sequentially satisfy $\phi_1$ and then $\phi_2$. The statement 4) says that the trajectory should satisfy either $\phi_1$ or $\phi_2$. We say a trajectory $\zeta$ satisfies specification $\phi$ if there is a time step $h$ such that the prefix $\zeta_{0:h}$ satisfies $\phi$.

In addition, every SPECTRL specification $\phi$ is guaranteed to have an equivalent directed acyclic graph (DAG), termed as abstract graph \citep{jothimurugan2019composable}. An abstract graph $\mathcal{G}$ is defined as $\mathcal{G}::=(Q,E,q_0,F,\kappa)$, where $Q$ is the set of nodes, $E\subseteq Q\times Q$ is the set of directed edges, $q_0\in Q$ denotes the initial node, $F\subseteq Q$ denotes the accepting nodes, subgoal region mapping $\beta: Q\to 2^{\mathcal{S}}$ which denotes the subgoal region for every node in $Q$, and safe trajectories $\mathcal{Z}_{\text{safe}}=\cap_{e\in E}\mathcal{Z}^e_{\text{safe}}$ where $\mathcal{Z}^e_{\text{safe}}$ denotes the safe trajectories for any edge $e\in E$. Note that the environmental MDP $\mathcal{M}$ is connected with task specification $\phi$ and $\mathcal{G}_{\phi}$ by $\beta$ and $\mathcal{Z}^e_{\text{safe}}$ which may change for different tasks.  Furthermore, the function $\kappa$ labels each edge $e:=q\to q'$ with the symbol $b_e$ (labeled edge denoted as $e:=q\xrightarrow{b_e} q'$). Given $\kappa$, the agent transits from node $q$ to $q'$ when the states $s_i$ and $s_{i+l}$ of trajectory $\zeta$ satisfy $s_i\in\beta(q)$ and $b_e\subseteq L(s_{i+l})$ for some $l\ge 0$.

Given a task specification $\phi$, the corresponding abstract graph $\mathcal{G}_{\phi}$ can be constructed based on its definition, such that, for any trajectory $\zeta\in\mathcal{Z}$, we have $\zeta\models\phi$ if and only if $\zeta\models{G}_{\phi}$. Hence, the RL problem for task $\phi$ can be equivalent to the reachability problem for $\mathcal{G}_{\phi}$. 
It is obvious that every task DAG has a single initial node in SPECTRL language, which can be converted into a tree.

\subsubsection{Sub-task Definition}
\label{sec:sub-task}
Given the DAG $\mathcal{G}_{\phi}$ corresponding to task specification $\phi$, we can define sub-tasks based on edges of the DAG. Formally, an edge from node $q$ to $p\in Q$ can define a reach-avoid sub-task specified by the following SPECTRL formula:
\begin{equation}
\small
    \text{Sub-Task}(q,p):=\text{achieve}(b_{(q,p)})\text{ ensuring } \bigg(\bigwedge_{r\in\mathcal{N}(q), r\neq p}\neg b_{(q,r)}\bigg) \label{subtask}
\end{equation}
where $b_{(q,p)}$ is the propositional formula labeled over the edge $(q,p)$ in the DAG, and $\mathcal{N}(q)$ is the set of neighboring nodes to which the out-going edges of $q$ point in the DAG. For instance, in Figure \ref{fig:imp_mot_exp2}, the propositional formula over the edge $(q_0, q_2)$ is $b_{(q_0,q_2)}=\neg d\wedge a$. When $e=(q,p)$, the notation Sub-Task$(e)$ is same as Sub-Task$(q,p)$ defined in \eqref{subtask}, e.g., Sub-Task$(q_0, q_2):=\text{achieve}(a)\text{ensuring}(\neg b\wedge\neg d)$ in Figure \ref{fig:imp_mot_exp2} after some algebra. 

For each Sub-Task$(q,p)$ and any MDP state $s_0\in\mathcal{S}$, there is a policy $\pi_{(q,p)}$ which can guide the agent to produce a trajectory $s_0s_1\ldots s_n$ in MDP. It induces the path $qqq\ldots qp$ in the DAG, meaning that the agent's DAG state remains at $q$ until it transits to $p$, i.e., $s_n\in\beta(p)$ and $s_i\notin\beta(p)$ for $i<n$. In this work, since we consider the dependencies of sub-tasks, the policy $\pi_{(q,p)}$ is also dependent on the future sub-tasks to complete.

Given the environmental MDP $M$, for any SPECTRL task specification $\phi$, the agent first transforms $\phi$ to its corresponding DAG (abstract graph) $\mathcal{G}_{\phi}=(Q,E,q_0,F,\beta,\mathcal{Z}_{\text{safe}},\kappa)$. Then, the sub-tasks of all the edges can be obtained from $\mathcal{G}_{\phi}$ based on the equation \eqref{subtask}. In this work, we can assume that for every edge sub-task of the DAG, the achieve part only has conjunction of propositions (denoted as $p_+$ for reaching) and the ensuring part only has conjunctions of negated propositions (denoted as $p_-$ for avoidance). For example, for the sub-task achieve$(b)$ensuring$(\neg a\wedge\neg d)$ (i.e., $b\wedge\neg a\wedge\neg d$), we have $p_+=\{b\}$ and $p_-=\{a,d\}$. Whenever the achieve part in \eqref{subtask} contains disjunction, this sub-task will decomposed further into sub-tasks in parallel edges, until every sub-task only achieves conjunction of propositions.

\subsection{Problem Formulation}
\label{sec:formulation}
Given the environmental MDP $M$ with unknown state transition dynamics, a SPECTRL specification $\phi$ represents the logic compositional task consisting of temporally extended sub-tasks, and $\mathcal{G}_{\phi}$ is the DAG (abstract graph) corresponding to the task $\phi$. 

The target of this work is to train an reinforcement learning (RL) agent in a data-efficient manner which can be zero-shot generalized to complete any unseen SPECTRL task $\phi$ without further training. In addition to task completion, we also consider the optimality of the found solution for the unseen task $\phi$, maximizing the accumulated environmental rewards during task completion. Specifically, the reward function of MDP $\mathcal{M}$ is unknown to the agent, and the reward of any state $s$ is available to the agent only whenever $s$ is visited.

\section{Methodology}
\label{sec:method}
In the following sections, we first detail the modules and operational mechanisms of the proposed framework. Next, we outline the training algorithm, describing the processes for training both low-level and high-level modules. We also propose specific training techniques aimed at enhancing robustness and data efficiency throughout the learning process, including curriculum, experience relabeling and proposition avoidance, which are introduced in Appendix \ref{sec:training_app}.

\subsection{Architecture}
\label{sec:architecture}
The proposed framework consists of high-level and low-level modules. 
The high-level module is essentially an implicit planner which selects the next sub-task for the low-level agent to complete. 
Based on the feature of future sub-tasks, the implicit planner is directly trained to predict the best selection of next sub-task and also estimate the return for completing the rest of task, which are passed to the low-level agent for guidance.
The low-level module is trained to achieve the assigned sub-task together with the estimated return which makes the low-level policy look into the future sub-tasks. This approach fastens the training of the low-level module, improving the learning efficiency in long-horizon tasks. 

\begin{figure*}
    \centering
    \fontsize{8pt}{8pt}\selectfont
    \def\svgwidth{6in}
    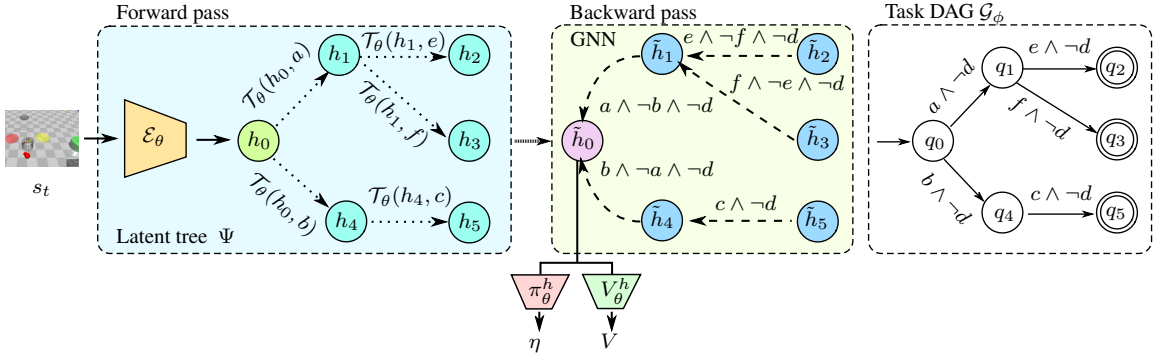
    \caption{\small Diagram of implicit planner as the high-level agent. The DAG (abstract graph) of the task is shown in the rightmost figure. The latent tree is spanned by the encoder $\mathcal{E}_{\theta}$ and latent transition model $\mathcal{T}_{\theta}$ in the forward pass, while the feature of future sub-tasks is extracted by GNN ($\mathcal{M}_{\theta}, \mathcal{U}_{\theta}$) in the backward pass. The sub-task $\eta$ and estimated return $V$ are predicted by policy $\pi_{\theta}^h$ and value networks $V^h_{\theta}$, respectively. Note that in GNN, every edge is labeled by a corresponding sub-task derived from the task DAG, and the feature of a edge is the binary encoding of positive and negative propositions of the labeled sub-task.}
    \label{fig:imp-plan-diagram}
\end{figure*}

\subsubsection{High-level Module}
\label{sec:high-agent}
When the dependencies among sub-tasks are accounted for, the planning problem of selecting the next high-level sub-task no longer adheres to the Markovian property. This limitation prevents the use of the commonly applied value iteration (VI) method for sub-task selection, as VI relies on Bellman equations \citep{sutton2018reinforcement} to compute the value function and is effective only when the Markovian property holds. To address this, we introduce an implicit planner that directly predicts the optimal next sub-task and estimates the expected return for completing the remaining task based on an embedding that represents future sub-tasks and states. This embedding is generated by a graph neural network (GNN) \citep{scarselli2008graph,zhou2020graph} and a latent transition model \citep{kipfcontrastive,van2020plannable}, described below.

As shown in Figure \ref{fig:imp-plan-diagram}, the proposed implicit planner consists of an encoder $\mathcal{E}_{\theta}$, a latent state transition model $\mathcal{T}_{\theta}$, a GNN $(\mathcal{M}_{\theta},\mathcal{U}_{\theta})$, a policy network $\pi^h_{\theta}$ and a value network $V^h_{\theta}$. All components of the implicit planner are trained together end-to-end, so their trainable parameters are collectively represented as $\theta$. The implicit planner operates through both forward and backward passes.

\noindent
{\bf Forward Pass.}
In the forward pass, the planner generates latent representations of the current and future states by using the encoder $\mathcal{E}_{\theta}$ and the latent dynamic model $\mathcal{T}_{\theta}$, iteratively constructing a tree $\Psi$ whose node features are predicted latent representations of states. Given the current environmental state $s_t$, the encoder first derives its latent representation, denoted as $h_0:=\mathcal{E}_{\theta}(s_t)$, which serves as the root of the latent tree $\Psi$. Following the structure of the task directed acyclic graph (DAG) $\mathcal{G}_{\phi}$, the tree $\Psi$ is expanded from $h_0$ until every accepting node in $F$ of the DAG $\mathcal{G}_{\phi}$ is included in $\Psi$.

Expanding $\Psi$ from a node $n$ entails adding all nodes in $\mathcal{G}_{\phi}$ connected through edges that originate from $n$. Specifically, the latent state of node $n$ (i.e., $h_n$) and the subgoals associated with its outgoing edges in $\mathcal{G}_{\phi}$ are input to the latent transition dynamics model $\mathcal{T}_{\theta}$, which then predicts the subsequent latent states. These predicted states serve as features for new nodes, which are added to $\Psi$ as the children of node $n$. This expansion process iterates until all subgoals in $\mathcal{G}_{\phi}$ are incorporated into $\Psi$. An example is illustrated in Figure \ref{fig:imp-plan-diagram}. It is worth noting that $\Psi$ is built based on the subgoals (positive propositions of sub-tasks) associated with edges in $\mathcal{G}_{\phi}$, rather than directly from sub-tasks.

\noindent
{\bf Backward Pass.}
In the backward pass, the GNN component operates over a graph modified from the latent tree $\Psi$ to extract an embedding vector, $\tilde{h}_0$, which represents the completion of the remaining task. Specifically, each edge in the latent tree $\Psi$ is reversed in direction and labeled with the sub-task, derived from corresponding edges in the task DAG $\mathcal{G}_{\phi}$ based on its definition in \eqref{subtask}, resulting in a new graph. A multi-layer GNN is then applied to this new graph to extract an embedding vector that encapsulates future sub-tasks in the remaining task, denoted by the node feature $\tilde{h}_0$. An example of this process is illustrated in Figure \ref{fig:imp-plan-diagram}. Finally, based on $\tilde{h}_0$, the policy and value function determine the selection of the next sub-task and estimate the return for the low-level agent.

It’s important to note that, in the backward pass, the sub-task on each edge of the graph is derived based on \eqref{subtask} and differs from the sub-task in the task DAG $\mathcal{G}_{\phi}$. This distinction is because the derived sub-task explicitly incorporates conditions to avoid accidentally completing any neighboring sub-tasks, which is not explicitly addressed in the original task DAG $\mathcal{G}_{\phi}$, as shown in the example of Figure \ref{fig:imp_mot_exp2}.

\noindent
{\bf Encoder.} The encoder function, $\mathcal{E}_{\theta}:\mathcal{S}\to\mathbb{R}^d$, takes an environmental state as input and outputs its latent representation. The encoder’s neural architecture is tailored to the environment: a CNN is employed for pixel-based environments, while an MLP is used for environments with continuous observations.

\noindent
{\bf Latent Transition Model.} The latent transition function, $\mathcal{T}_{\theta}:\mathbb{R}^{d}\times\mathcal{P}\to\mathbb{R}^d$, predicts the next latent state by using the latent of the current state and the subgoals (positive propositions of the sub-task). Given the current state $s$ and sub-task $\eta$, the next latent state is predicted as $\mathcal{E}_{\theta}(s) + \mathcal{T}_{\theta}(\mathcal{E}_{\theta}(s), p^{\eta}+)$, where $p^{\eta}+$ is the binary encoding of the subgoals of sub-task ${\eta}$. This function models the changes in the latent state caused by completing a sub-task. In implementation, $\mathcal{T}_{\theta}$ is typically realized using an MLP.

\noindent
{\bf Graph Neural Network.} The GNN is employed to generate an embedding that represents the progress toward completing the remaining tasks. For each node $k$ in the GNN, it first gathers a set of incoming messages from each connected node $j$ with an edge $(j, k)$ directed from $j$ to $k$. This is achieved using the message-passing function $\mathcal{M}_{\theta}$, which takes as input the features of nodes $k$ and $j$ ($\tilde{h}_k$ and $\tilde{h}_j$) along with the edge feature $e{(j, k)}$. The initial feature of each node is its latent state $h_k$, predicted by $\mathcal{T}_{\theta}$ during the forward pass, and it is subsequently updated with the incoming messages using the update function $\mathcal{U}_{\theta}$.

The edge feature $e_{(j, k)}$ is a binary encoding of the sub-task $b_{(k, j)}$ associated with the edge $(j, k)$. Specifically, this feature is created by concatenating two binary vectors that separately represent the positive and negative propositions of the sub-task.

In every layer of GNN, the incoming messages of node $k$ are first aggregated by summation as below,
\begin{equation}
    m_{k} = \bigoplus_{j\in\mathcal{N}(\tilde{h})}\mathcal{M}_{\theta}(\tilde{h}_j, \tilde{h}_k, e_{(j, k)}) \label{message}
\end{equation}
Then, the node feature $\tilde{h}_k$ is updated with incoming message, i.e., $\tilde{h}_k\leftarrow \mathcal{U}_{\theta}(\tilde{h}_k,m_k)$. 
In implementation, both $\mathcal{M}_{\theta}$ and $\mathcal{U}_{\theta}$ are realized by MLPs.

Since the direction of each edge in $\Psi$ is reversed in the graph used for the backward pass, multiple iterations of applying the functions $\mathcal{M}_{\theta}$ and $\mathcal{U}_{\theta}$ in the GNN enable information from each future sub-task to back-propagate to the root node. Consequently, the root node feature, denoted as $\tilde{h}_0$, encapsulates the status of completing future sub-tasks within the remaining task.

\noindent
{\bf Policy and value function.} The policy function $\pi^h_{\theta}:\mathbb{R}^d\to[0,1]^{|\mathcal{P}|}$ maps the embedding $\tilde{h}_0$ extracted by GNN to a distribution of {\it feasible} next sub-tasks. The next sub-task assigned to the low-level agent is sampled from this output distribution. 
The value function $V^h_{\theta}:\mathbb{R}^d\to\mathbb{R}$ maps the embedding $\tilde{h}_0$ to the estimated return for completing the remaining task starting from the current state.

\noindent
{\bf Remark.} 
Previous approaches to planning for logic-based compositional tasks have utilized value iteration \citep{araki2019learning,araki2021logical}, Dijkstra's algorithm \citep{he2015towards}, and heuristic-based search algorithms \citep{khalidi2020t, gujarathi2022mt}. These methods rely on the assumption that the cumulative rewards for completing each sub-task are independent of others, meaning the Markovian property must hold. 
However, as illustrated in the examples in Section \ref{sec:intro}, this work considers dependencies between sub-tasks, where the cumulative rewards for completing one sub-task depend on future sub-tasks, breaking the Markovian property and making standard planning algorithms inapplicable. To address this, we leverage the generalization capabilities of GNNs to train a value function for sub-tasks and states through supervised learning, enabling the high-level policy to be trained using the PPO algorithm \citep{schulman2017proximal}.

\subsubsection{Low-level Module}
\label{sec:low-agent}
\begin{figure}
    \centering
    \fontsize{10pt}{10pt}\selectfont
    \def\svgwidth{2.45in}
    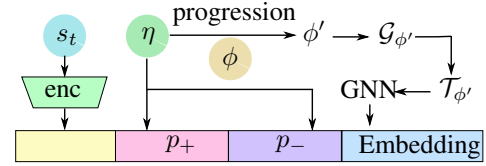
    \caption{\small Diagram of processing inputs to the low-level policy $\pi^l_{\omega}$ and value function $V^l_{\omega}$. $\eta$ is the sub-task assigned by the high-level module. $\phi$ is the target task to complete. $\phi'$ is the progression of $\phi$ with $\eta$. The embedding is the representation of $\phi'$ produced by the GNN. $s_t$ is the environmental observation. "Enc" is the encoder mapping observation into a latent vector.}
    \label{fig:low-agent-inputs}
\end{figure}

The target of low-level module is to complete the sub-task $\eta$ specified by the high-level module. As discussed in the Sub-task Definition of Section \ref{sec:ltl}, every sub-task in SPECTRL language is a reach-avoid task, and can be decomposed into positive proposition (to achieve) and negative propositions (to avoid), denoted as $p_+$ and $p_-$, respectively. Hence, the low-level policy and value functions, denoted as $\pi^l_{\omega}$ and $V^l_{\omega}$, are conditioned on $p_+$ and $p_-$ encoded into binary vectors. The diagram of processing inputs to the low-level agent is in Figure \ref{fig:low-agent-inputs}.

To account for dependencies of future sub-tasks, both $\pi^l_{\omega}$ and $V^l_{\omega}$ are conditioned on the DAG of the remaining task $\phi'$, denoted as $\mathcal{G}_{\phi'}$. Here, $\phi'$ represents the remainder of the task $\phi$ after completing sub-task $\eta$. Essentially, $\phi'$ is the progression \citep{kvarnstrom2000talplanner,vaezipoor2021ltl2action} of task $\phi$ once sub-task $\eta$ is achieved. For example, in the task shown in Figure \ref{fig:imp_mot_exp2}, if sub-task $b\wedge\neg a\wedge\neg d$ (on the edge from state 0 to 1) is completed, the progression of the task becomes $c\wedge\neg d$, indicating the remaining part of the task to be accomplished.

\noindent
{\bf DAG Processing.} However, the functions $\pi^l_{\omega}$ and $V^l_{\omega}$ cannot directly process a DAG. To address this, we propose using a GNN to generate an embedding from the tree representation of the DAG $\mathcal{G}_{\phi'}$, which can be directly used as input to $\pi^l_{\omega}$ and $V^l_{\omega}$. Since any DAG with a single source node can be equivalently converted into a tree, we first convert the DAG of $\phi'$, denoted $\mathcal{G}_{\phi'}$, into its corresponding tree $\mathcal{T}_{\phi'}$. 

This tree $\mathcal{T}_{\phi'}$ differs from the high-level latent tree $\Psi$. Given that the low-level policy operates at every time step, it’s essential to keep the low-level module as streamlined as possible. Thus, the latent transition model is omitted, and each node’s initial feature is set to an all-zero vector. Additionally, we ignore the negative propositions of future sub-tasks, using only the binary encoding of the positive propositions (subgoals) as the edge feature. However, the negative propositions of the current sub-task $\eta$ are still encoded as a binary vector $p_-$, as shown in Figure \ref{fig:low-agent-inputs}. 

With these defining node and edge features, a multi-layer GNN is applied to $\mathcal{T}{\phi'}$, with the direction of each edge reversed. The embedding at the root node then represents the characteristics of all future subgoals to accomplish. The process of deriving $p_+$, $p_-$, and the embedding for future subgoals is illustrated in Figure \ref{fig:low-agent-inputs}. Note that this low-level GNN differs from the one used at the high level and is trained alongside other low-level components.

\subsection{Algorithm}
\label{sec:algo}
The high-level and low-level modules are trained independently. In the high level, all components are trained end-to-end to predict the optimal selection of the next sub-task to complete and to estimate the return for completing the remainder of the task. In the low level, the module is trained to accomplish the assigned sub-task while accounting for dependencies from future sub-tasks. To enhance the sample efficiency of the learning process, several training techniques are introduced, including a training curriculum, experience replay, and proposition avoidance.

\noindent
{\bf High-level Transition Data.} 
We define a high-level transition tuple as $(s, \phi, \eta, R, s', \phi')$, which indicates that starting from state $s$, the agent completes the sub-task $\eta$ and reaches state $s'$. Here, $\phi'$ is the progression of task $\phi$ after sub-task $\eta$ is completed, representing the remaining part of the task, and $R$ is the accumulated discounted reward earned while completing $\eta$.
The high-level transition buffer $\Gamma^h$ stores these transition tuples collected from all trajectories, but only includes transitions where completing the sub-task $\eta$ resulted in progression over $\phi$ (i.e., $\phi' \neq \phi$).
For a trajectory $\zeta = {(s_l, a_l, r_l)}_{l=0}^{H-1}$, where $K_{\zeta}$ sub-tasks are completed sequentially, we denote these sub-tasks as $\eta_0, \eta_1, \ldots, \eta_{K_{\zeta}-1}$, with the time steps at which they are completed as $t_0, t_1, \ldots, t_{K_{\zeta}-1}$. The accumulated rewards obtained for completing each sub-task $\eta_i$ are defined as $R_i := \sum_{\tau=t_{i-1}}^{t_i} \gamma^{\tau - t_{i-1}} r_{\tau}$, where $r_{\tau}$ represents the environmental reward at time step $\tau$.

\subsubsection{High-level Training}
\label{sec:high-train}
The effectiveness of the high-level module depends on the embedding vector that represents the remaining task, which is extracted by the latent transition model and GNN. Therefore, it is essential to train the encoder, latent transition model, and GNN effectively. Drawing inspiration from previous work on learning latent dynamic spaces \citep{bordes2013translating,kipfcontrastive}, we utilize the TransE \citep{kipfcontrastive} loss to train the encoder $\mathcal{E}_{\theta}$ and the latent transition function $\mathcal{T}_{\theta}$ together. For any high-level transition data $(s,\phi,\eta,R,s',\phi')$ and a negatively sampled state $\tilde{s}$, the TransE loss can be expressed as below:
\begin{eqnarray}
    \mathcal{L}_{\text{TransE}}((s,\eta,s'),\tilde{s};\theta)
    &=&d(\mathcal{E}_{\theta}(s)+\mathcal{T}_{\theta}(\mathcal{E}_{\theta}(s),\eta), \mathcal{E}_{\theta}(s')) \nonumber \\
    &&+ \max(0, \xi-d(\mathcal{E}_{\theta}(s), \mathcal{E}_{\theta}(\tilde{s}))) \label{transe}
\end{eqnarray}
where $\theta$ are the trainable parameters, $d$ is the distance function which is chosen as the Euclidean distance in this work, and $\xi$ is a positive hyper-parameter. The tasks $\phi,\phi'$ and reward $R$ are not used in the training loss of $\mathcal{E}_{\theta}$ and $\mathcal{T}_{\theta}$.

For the GNN part, models $\mathcal{M}_{\theta}$ and $\mathcal{U}_{\theta}$ are trained together with policy $\pi^h_{\theta}$ and value networks $V^h_{\theta}$ in an end-to-end manner. Since the training curriculum is designed to start from simple tasks, there is no need to pre-train the GNN part. 

The components of the high-level module are jointly trained using the PPO algorithm \citep{schulman2017proximal} with a set of feasible sub-tasks serving as the action space. Based on the high-level transition buffer $\Gamma^h$, the PPO loss is calculated by evaluating the outputs of the policy and value networks through the forward and backward passes, as detailed in Section \ref{sec:high-agent}.
One iteration of training the high-level module can be summarized as the following steps:
\begin{enumerate}
    \item Sample trajectories $\zeta$ from the replay buffer $\mathcal{B}$ which forms the high-level transition dataset $\Gamma^h$;
    \item Based on transition tuples in $\Gamma^h$, compute the PPO \citep{schulman2017proximal} and TransE \eqref{transe} losses, where the negative samples $\tilde{s}$ in \eqref{transe} are randomly sampled from states in $\Gamma^h$;
    \item Update parameters $\theta$ of all the components in the high-level module together, with gradients of the following loss function:
    \begin{eqnarray}
        \lefteqn{\mathcal{L}(\Gamma^h;\theta)=\mathcal{L}_{\text{PPO}}(\Gamma^h;\theta)} \nonumber \\
        &&+\lambda \sum_i \mathcal{L}_{\text{TransE}}((s_{t_i}, \eta_i, s_{t_{i+1}}),\tilde{s}_i;\theta) \label{high-loss}
    \end{eqnarray}
    where $\lambda$ is a hyper-parameter to balance two loss terms, chosen as $0.01$ in this work.
\end{enumerate}
Note that since $V^h_{\theta}$ is trained with $R_i$ in the value loss of PPO, where $R_i$ is a discounted accumulated rewards in multiple steps and hence $V^h_{\theta}$ is essentially updated by a multi-step Bellman operator (BO) \citep{sutton2018reinforcement}. 

\subsubsection{Low-level Training}
\label{sec:low-train}
The low-level module is trained using the PPO algorithm \citep{schulman2017proximal} with transition tuples from sampled trajectories, where the target sub-tasks and estimated returns are generated by the high-level module. As described in Section \ref{sec:low-agent}, the low-level agent is optimized to complete the assigned sub-task $\eta_i$ with maximum accumulated environmental rewards. As shown in Figure \ref{fig:low-agent-inputs}, the inputs to the low-level policy $\pi^l_{\omega}$ are derived from the assigned sub-task ($\eta_i$) and the remaining task to complete ($\phi_i$). Upon completion of $\eta_i$ at time step $t_i$, the task $\phi_i$ progresses to $\phi_{i+1}$ based on the completion of $\eta_i$ (the {\it progression} is described in Section \ref{sec:low-agent}), and the high-level module then assigns the next sub-task, $\eta_{i+1}$. At regular intervals, and until an assigned sub-task is completed, the low-level module (comprising policy $\pi^l_{\omega}$ and value function $V^l_{\omega}$) is trained using PPO with low-level transition data directly obtained from the collected trajectories.

Because the low-level module is conditioned on the remaining task ($\phi_i$), which includes future sub-tasks, $\pi^l_{\omega}$ and $V^l_{\omega}$ need to be trained with reward information that considers the completion of these future sub-tasks. To address this, we introduce the {\it max operator} to incorporate the estimated returns of future sub-tasks into the target computation for training $V^l_{\omega}$. Specifically, whenever a sub-task $\eta_i$ is completed (i.e., at time step $t_i$), $V^l_{\omega}$ is trained to predict the maximum between the one-step return generated by $V^l_{\omega}$ itself and the multi-step return $V^h_{\theta}$ estimated by the high-level module.
Therefore, at time step $t_i$, the target values to train $V^l_{\omega}$ can be written as 
\begin{eqnarray}
    \lefteqn{V^l_{\text{target}}(s_{t_i},p^{i+1}_+,p^{i+1}_-,\phi'_{i+1})} \nonumber \\
    &:=&\max\{r_{t_i} + \gamma V^l_{\omega}(s_{t_i+1},p^{i+1}_+,p^{i+1}_-,\phi'_{i+1}), \nonumber \\
    &&V^h_{\theta}(s_{t_i}, \phi_{i+1})\} \label{low-v-max}
\end{eqnarray}
where $p^{i+1}_+$ and $p^{i+1}_-$ are positive (subgoals) and negative propositions (avoidance) decomposed from $\eta_{i+1}$, and $\phi_{i+1}'$ is the progression of $\phi_{i+1}$ with $\eta_{i+1}$. Since $\pi^l_{\omega}$ is trained alongside $V^l_{\omega}$ using PPO, the low-level policies can account for dependencies among future sub-tasks, rather than focusing solely on completing the current sub-task ($\eta_i$). This approach enhances the global optimality of task completion by promoting strategies that consider the overall task dependencies.

We have two main intuitions for using the max operator in \eqref{low-v-max}. First, since $V^h_{\theta}$ is trained with a multi-step Bellman operator (BO) at the high level, while $V^l_{\omega}$ is trained with a regular single-step BO at the low level, reward information for future sub-tasks can be propagated back to the current sub-task $\eta_i$ much faster at the high level than at the low level. Thus, in the early stages of learning, value estimates from $V^h_{\theta}$ are generally higher than those from $V^l_{\omega}$. The max operator can then leverage $V^h_{\theta}$'s estimated returns to accelerate learning in the low-level module, addressing the sparse reward challenge in long-horizon tasks. Second, in the later stages of training, if $V^h_{\theta}$ is occasionally influenced by sub-optimal trajectories and yields lower value estimates, the max operator allows $V^l_{\omega}$ to disregard these sub-optimal values from $V^h_{\theta}$. This prevents the low-level module from deviating from its optimal trajectory, maintaining its effectiveness.

\section{Experiments}
\label{sec:experiments}
Our experiments aim to evaluate the performance of a multi-task RL agent trained using the proposed framework, focusing on learning efficiency, optimality, and generalization. Specifically, the section on overall performance examines whether the proposed framework outperforms baselines in terms of optimality and learning efficiency when sub-task dependencies are present. Next, ablation studies investigate the impact of considering future sub-tasks within the low-level module and assess the contribution of experience relabeling to learning efficiency. Finally, we visualize several trajectories of trained agents in the Walk domain to illustrate the framework’s effectiveness in achieving optimality and avoiding unintended subgoals.

Before presenting the experiment results, we will first introduce the environments. Then the training setup and baseline methods will be presented. Finally, the experiments about overall performance comparison will be demonstrated. Other experiment results and algorithmic details are deferred to Appendix.

\begin{figure}
    \centering
    \subfigure[Letter]{
        \fontsize{8pt}{10pt}\selectfont
        \def\svgwidth{.8in}
        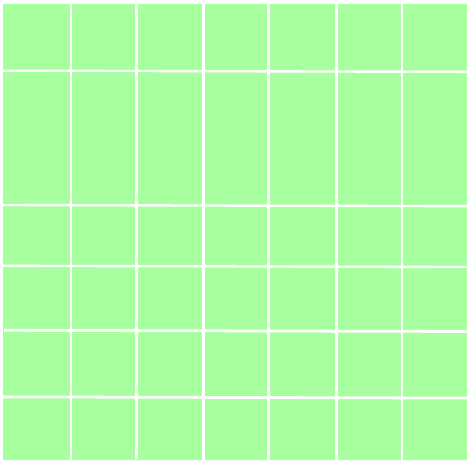
        \label{fig:letter_map}
    }
    \subfigure[Walk]{
        \centering
        \fontsize{8pt}{10pt}\selectfont
        \def\svgscale{.9}
\begingroup%
  \makeatletter%
  \providecommand\color[2][]{%
    \errmessage{(Inkscape) Color is used for the text in Inkscape, but the package 'color.sty' is not loaded}%
    \renewcommand\color[2][]{}%
  }%
  \providecommand\transparent[1]{%
    \errmessage{(Inkscape) Transparency is used (non-zero) for the text in Inkscape, but the package 'transparent.sty' is not loaded}%
    \renewcommand\transparent[1]{}%
  }%
  \providecommand\rotatebox[2]{#2}%
  \newcommand*\fsize{\dimexpr\f@size pt\relax}%
  \newcommand*\lineheight[1]{\fontsize{\fsize}{#1\fsize}\selectfont}%
  \ifx\svgwidth\undefined%
    \setlength{\unitlength}{82.11122167bp}%
    \ifx\svgscale\undefined%
      \relax%
    \else%
      \setlength{\unitlength}{\unitlength * \real{\svgscale}}%
    \fi%
  \else%
    \setlength{\unitlength}{\svgwidth}%
  \fi%
  \global\let\svgwidth\undefined%
  \global\let\svgscale\undefined%
  \makeatother%
  \begin{picture}(1,0.76604922)%
    \lineheight{1}%
    \setlength\tabcolsep{0pt}%
    \put(0,0){\includegraphics[width=\unitlength,page=1]{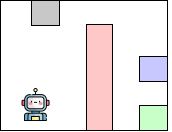}}%
    \put(0.22607662,0.68000773){\color[rgb]{0,0,0}\makebox(0,0)[lt]{\lineheight{1.25}\smash{\begin{tabular}[t]{l}$d$\end{tabular}}}}%
    \put(0,0){\includegraphics[width=\unitlength,page=2]{walk_env.pdf}}%
    \put(0.13053132,0.37604524){\color[rgb]{0,0,0}\makebox(0,0)[lt]{\lineheight{1.25}\smash{\begin{tabular}[t]{l}$e$\end{tabular}}}}%
    \put(0.55352648,0.28416148){\color[rgb]{0,0,0}\makebox(0,0)[lt]{\lineheight{1.25}\smash{\begin{tabular}[t]{l}$a$\end{tabular}}}}%
    \put(0.86662965,0.34645557){\color[rgb]{0,0,0}\makebox(0,0)[lt]{\lineheight{1.25}\smash{\begin{tabular}[t]{l}$b$\end{tabular}}}}%
    \put(0.87029304,0.05122367){\color[rgb]{0,0,0}\makebox(0,0)[lt]{\lineheight{1.25}\smash{\begin{tabular}[t]{l}$c$\end{tabular}}}}%
  \end{picture}%
\endgroup%

        \label{fig:walk_env}
    }
    \subfigure[Zone]{
        \centering
        \includegraphics[width=.8in, height=0.8in]{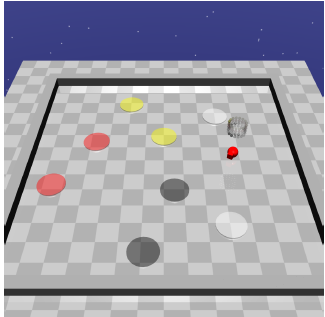}
        \label{fig:robot_zone}
    }
    \caption{Environments}
    \label{fig:envs}
\end{figure}
\subsection{Environments}
\label{sec:exp-setup}
We conducted experiments in various environments featuring both discrete and continuous action and state spaces. Each environment is procedurally generated, with object layouts and positions randomized upon reset. The agent does not know the positions or properties of objects in advance, making it impossible to solve these environments using simple tabular-based methods. Each task is defined by a SPECTRL specification, expressed through symbolic propositions given by the labeling function (introduced in Section \ref{sec:rl}). The agent's goal is to complete the specified task while maximizing accumulated rewards. The example screen shots of environments are shown in Figure \ref{fig:envs}. The detailed introduction of environments are in Appendix \ref{sec:app_envs}.

\noindent
{\bf Sub-task Dependencies.}
In the experiments, sub-task dependencies may arise from many sources, including avoidance requirements, multiple instances of symbols, and the reward function of subgoals. First, completing one sub-task might require avoiding the subgoals of another, creating dependencies between them. {\it Note that the sub-task mentioned here is defined by \eqref{subtask} based on the task DAG, which may not be directly expressed on the task DAG.} For instance, in Figure \ref{fig:imp_mot_exp2}, if the agent's optimal plan is to first reach automaton state 1 and then 2, the sub-task $b\wedge\neg a\wedge\neg d$ depends on $a\wedge\neg b\wedge\neg d$. This is because reaching subgoal "a" requires avoiding "b"; otherwise, the agent risks completing the wrong sub-task and transitioning to an unintended automaton state. Second, the presence of multiple instances of a symbol (e.g., several cells with the same letter in Figure \ref{fig:letter_map}) can make the low-level policy for one sub-task dependent on future sub-tasks, as the agent must decide which instance to visit for global optimality. Third, the reward function can also introduce dependencies; for example, in the Walk domain shown in Figure \ref{fig:walk_env}, visiting "b" before "c" may yield a higher reward than going directly to "c", correlating the sub-tasks "b" and "c". The first two types of dependencies are at a low level, while the third type operates at a higher (sub-task) level.

\begin{figure*}[!t]
    \centering
    \subfigure[Letter, First Set]{
        \centering
        \includegraphics[width=1.35in]{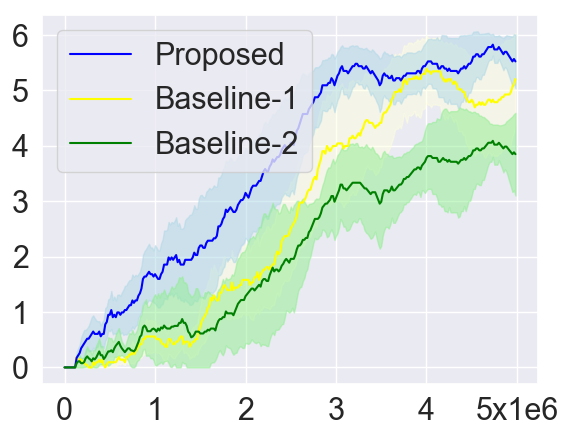}
        \label{fig:perf_letter}
    }
    \subfigure[Walk, First Set]{
        \centering
        \includegraphics[width=1.35in]{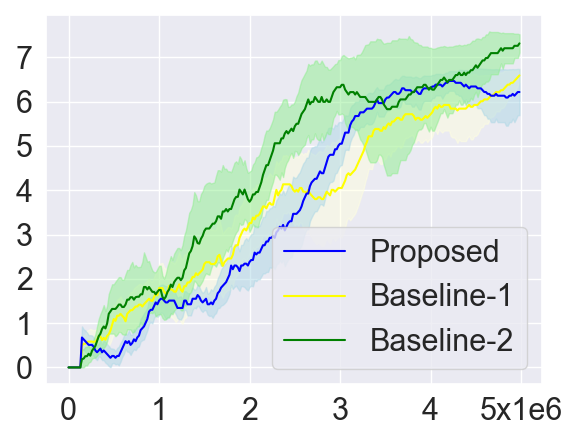}
        \label{fig:perf_walk}
    }
    \subfigure[Zone, First Set]{
        \centering
        \includegraphics[width=1.35in]{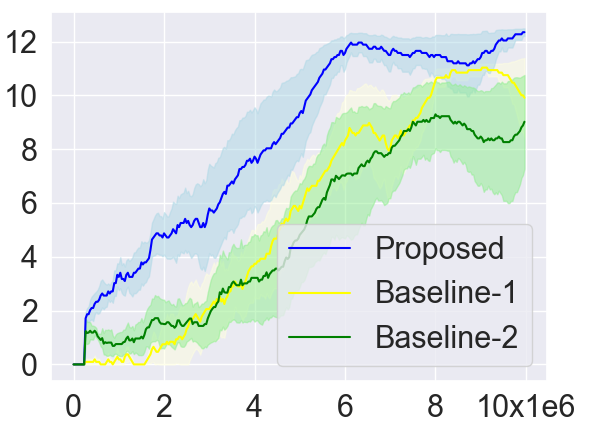}
        \label{fig:perf_zone}
    }

    \subfigure[Letter, Second Set]{
        \centering
        \includegraphics[width=1.35in]{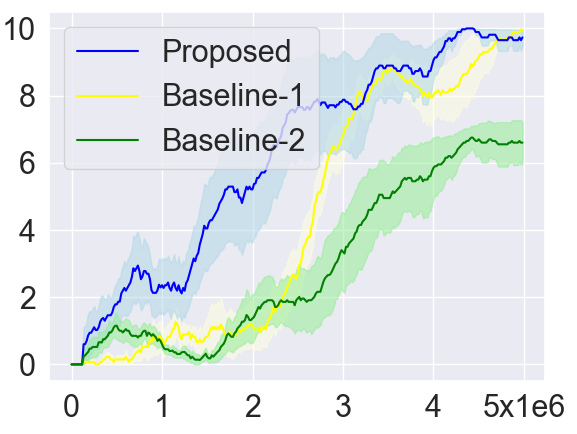}
        \label{fig:perf_letter1}
    }
    \subfigure[Walk, Second Set]{
        \centering
        \includegraphics[width=1.35in]{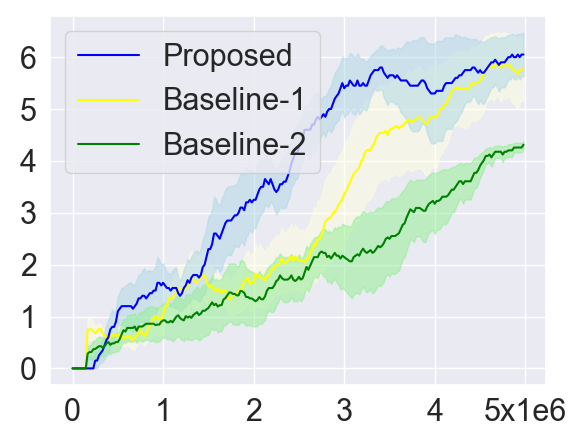}
        \label{fig:perf_walk1}
    }
    \subfigure[Zone, Second Set]{
        \centering
        \includegraphics[width=1.35in]{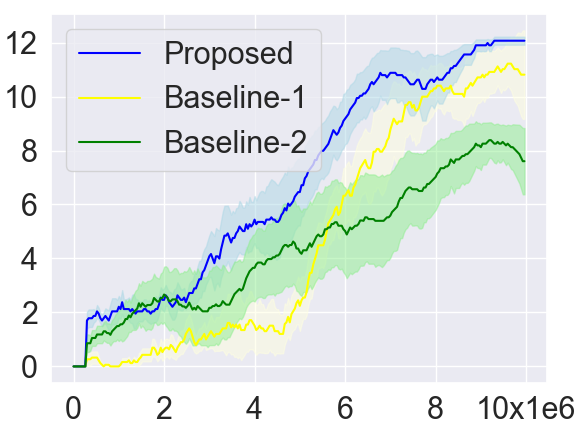}
        \label{fig:perf_zone1}
    }
    \caption{\small Performance Comparisons. The x-axis is the environmental step, and the y-axis is the average episodic return. In the first set of experiments, the reward is only given at the completion of the given task. In the second set, the reward of achieving a subgoal is dependent other subgoals achieved previously.}
    \label{fig:comparison}
    \vspace*{-10pt}
\end{figure*}

\subsection{Setup and Baselines}
\label{sec:training}
Training follows the curriculum described in Section \ref{sec:curriculum}. Every $5 \times 10^4$ environmental steps, the trained agent is evaluated on 10 randomly generated testing tasks over 100 episodes. These testing tasks are intentionally kept unseen during training. The evaluation metric is the average sum of rewards accumulated in each episode. The generation of testing task is presented in Appendix \ref{sec:hyper}. In both baselines, experience relabeling and training curriculum are also used for the fairness of comparison.

\noindent
{\bf Baseline-1.} We begin by comparing the proposed framework with LTL2Action \citep{vaezipoor2021ltl2action}, a leading RL algorithm for generalizing LTL satisfaction. LTL2Action leverages a graph convolutional network (GCN) to encode LTL task formulas, enabling policy decision-making to be conditioned on the entire task, allowing it to achieve the global optimality. However, unlike our approach, it does not decompose tasks into sub-tasks or employ a hierarchical framework, which results in a slower convergence rate. For implementation, we adapted its GCN to our task DAG structure, where for every edge $e$, the edge feature is the binary encoding of positive and negative propositions relevant to the sub-task associated with edge $e$. The node feature is updated with regular massage passing functions same as those in LTL2Action. 

\noindent
{\bf Baseline-2.} We then adapt the logical option framework \citep{araki2021logical} as the second baseline, referred to as "Option." Similar to our proposed framework, this approach addresses LTL tasks hierarchically. At the high level, it uses logic value iteration \citep{araki2019learning} to determine an optimal sequence of sub-tasks to complete the overall task, which is then assigned to the low level for execution. The accumulated rewards is estimated empirically for each individual subgoal separately. At the low level, the agent follows pre-trained policies to achieve each subgoal in the assigned sequence. Unlike our framework, this baseline does not account for future sub-task dependencies at either the high or low levels. Specifically, its high-level selection of sub-tasks overlooks reward-based dependencies among subgoals, and its low-level behavior is not trained to achieve global optimality by considering future sub-tasks.

\subsection{Overall Performance}
\label{sec:results}
In this section, we present the overall performance comparisons in terms of average return for completing testing tasks. Other experiments are included in Appendix. In each plot of the proposed framework, the x-axis is the number of environment steps used for training, while the y-axis is the average episodic return of executing testing tasks by the trained agent. 

In Figure \ref{fig:comparison}, the proposed method is compared with two baselines introduced in Section \ref{sec:training}. In the first set of experiments, the trained agent is evaluated on completing testing tasks with the minimum number of steps. Here, the reward is only provided upon completion of the task, and the agent must avoid any propositions specified in the safety condition. This set of experiments focuses on low-level sub-task dependencies, specifically requiring the agent to avoid subgoals of other sub-tasks and to reach designated subgoals while accounting for future subgoals.
In the second set of experiments, in addition to task completion, rewards are granted for achieving each subgoal. Additionally, the reward for certain subgoals depends on previously completed subgoals within the same episode, introducing high-level sub-task dependencies. Detailed information on reward dependencies among subgoals is provided in the Appendix \ref{sec:high-reward}.

In Figure \ref{fig:comparison}, we observe that the proposed framework outperforms Baseline-1 in terms of learning efficiency. Unlike the proposed method, Baseline-1 relies solely on a task-conditioned policy and does not decompose the task to leverage its compositional structure, which hinders its learning efficiency. The faster convergence of the proposed framework results from the multi-step return estimation performed by the high-level module, which backpropagates rewards for each sub-task and provides return estimates to the low-level module upon the completion of each sub-task. In contrast, Baseline-1 propagates rewards step-by-step, resulting in slower learning convergence of its value function.

As shown in Figure \ref{fig:comparison}, the proposed method outperforms Baseline-2 in terms of both average return and learning efficiency. In Baseline-2, the low-level policy is specifically trained for each assigned sub-task, aiming to reach positive propositions and avoid negative propositions relevant to that sub-task, without regard to other sub-tasks. This lack of awareness of dependencies compromises global optimality in low-level behavior, as shown in the example of Figure \ref{fig:reply_fig}. Additionally, when subgoal rewards depend on one another, Baseline-2's high-level planning cannot optimally select sub-tasks, as conventional planning methods cannot handle non-Markovian rewards. These limitations explain Baseline-2's lower average return.
However, in Figure \ref{fig:perf_walk}, Baseline-2 outperforms other methods, since the environment for this experiment is relatively simple and does not have too much sub-task dependencies. In this case, the simplicity of model architecture can make Baseline-2 perform best.

\section{Conclusion}
In this work, we propose a new hierarchical framework for generalizing compositional tasks in the SPECTRL language. 
In the high level, we propose to use an implicit planner to select next sub-task to complete for the low-level agent. In the low level, the policy is conditioned on both the assigned sub-task and the remaining task to complete. With comprehensive experiments, we demonstrate the advantages of the proposed framework over baselines in terms of optimality and learning efficiency. 
In the future, we plan to investigate the generalization of tasks specified by probabilistic logic language, and try to deploy proposed methods into real robots.


\bibliography{main}

\appendix

\newpage
\onecolumn

\section{Related Work}
\label{sec:related}
Applying the RL paradigm to solve logic compositional tasks has been explored in numerous prior studies. These methods typically start by converting the compositional task formula into an equivalent automaton representation and then create a product MDP by combining the environmental MDP with the task automaton. Prominent approaches utilizing product MDPs include Q-learning for reward machines (Q-RM) \citep{camacho2019ltl,icarte2018using,icarte2022reward}, LPOPL \citep{toro2018teaching}, and geometric LTL (G-LTL) \citep{littman2017environment}. Additionally, \citep{jothimurugan2021compositional} introduced the DiRL framework, which uses hierarchical RL to accomplish LTL tasks by integrating graph-based planning on the automaton to guide exploration for task satisfaction. However, these approaches are tailored to specific task formulas, requiring policies to be retrained from scratch for each new task. Thus, they lack zero-shot generalization.

Previous approaches have sought to train reusable skills or options to facilitate generalization in compositional task settings \citep{andreas2017modular,araki2021logical,leon2020systematic,leon2021nutshell}. In these methods, agents fulfill unseen tasks by sequentially combining pre-trained option policies through value iteration over potential subgoal choices, achieving satisfying success rate. However, they overlook inter-dependencies between subgoals, which can lead to suboptimal solutions, as demonstrated in Figures \ref{fig:reply_fig} and \ref{fig:imp_mot_exp2}. This limitation compromises both the optimality and sometimes even the feasibility of the solutions. Although \citep{kvarnstrom2000talplanner} considers causal dependencies among sub-tasks, it requires these dependencies to be explicitly provided. In contrast, our framework does not require the agent to know sub-task dependencies beforehand, making it applicable to scenarios with general and implicit sub-task dependencies.

In \citep{kuo2020encoding,vaezipoor2021ltl2action}, the authors propose task-conditioned policies to enable zero-shot generalization in compositional tasks by conditioning the policy on task embeddings extracted through recurrent graph neural networks \citep{kuo2020encoding} or graph convolutional networks \citep{vaezipoor2021ltl2action}. Although these methods can achieve optimal policies with extensive training, they struggle with poor learning efficiency. This inefficiency arises because they do not decompose tasks or leverage the inherent compositional structure of these tasks. In this work, we introduce a hierarchical RL framework for zero-shot generalization that achieves both global optimality and efficient learning, addressing these limitations.

Goal-conditioned reinforcement learning (GCRL) has long focused on training a unified policy for reaching arbitrary single goals within a specified goal space \citep{liu2022goal}. However, GCRL typically addresses scenarios where agents need to reach only a single goal per episode. In contrast, the compositional tasks in our work require achieving multiple subgoals under specific temporal order constraints. While some GCRL approaches introduce hierarchical frameworks that generate multiple subgoals within an episode \citep{li2021active, chane2021goal}, these frameworks primarily aid exploration, with subgoals achieved in any order. This lack of temporal constraints makes these GCRL methods incompatible with our setting, so that these methods are not compared against the proposed framework.

There is a recent work \citep{xu2024generalization} proposing to use future-dependent options to improve the optimality and learning efficiency of the generalization of compositional tasks. It is a similar work, but it does not consider safety issue and avoiding negative propositions of sub-tasks. This work proposes a new architecture and trains the agent to avoid negative propositions for the safety guarantee.

\section{Training Techniques}
\label{sec:training_app}
\subsection{Training Curriculum}
\label{sec:curriculum}
Due to the compositional nature of SPECTRL specifications, the number of possible task DAGs can be exceedingly large. Randomly sampling training tasks from this vast task space can lead to abrupt shifts in task difficulty, causing the agent to forget previously learned behaviors and significantly hindering training efficiency. To address this, we propose a curriculum-based approach where training task complexity gradually increases, adapting to the real-time performance of the agent, thereby enhancing training efficiency.

The training curriculum is specifically designed based on the number of sub-tasks required to complete a given task, with this number capped at $K$, a user-defined upper limit on task complexity. As described in Sections \ref{sec:high-agent} and \ref{sec:low-agent}, each task specification is structured as a tree within both the high-level and low-level modules, where each path (from root to leaf) represents a sequence of sub-tasks necessary to complete the task. Building on this, the curriculum at level $k$ ($\le K$) presents training tasks as sequences of $k$ randomly generated sub-tasks, with both positive and negative propositions chosen randomly within each sub-task. When the agent achieves a performance criterion--such as a success rate of 95\% for task completion--the training progresses to the next level, $k+1$, in the curriculum.

\subsection{Experience Replay}
\label{sec:replay}
In order to improve the sample efficiency of the learning process, we propose an experience replay method for generating more data to train the high-level and low-level modules. 
As introduced in Section \ref{sec:high-train}, the training data for the high-level module are transition tuples across consecutive sub-tasks, which are relatively scarce and more difficult to collect than the low-level training data. 
Besides, both high-level and low-level modules need to learn to process different tasks in the form of trees. 
So, trajectories collected for completing sequential tasks in the training curriculum are not enough. 
Therefore, we propose to relabel every collected trajectory with randomly generated tasks in the form of trees, which can generate more trajectory data to train the agent without increasing sample complexity. It is a modification of previous experience replay methods \citep{mnih2013playing,mnih2015human,voloshin2023eventual} in RL.

\begin{figure}
    \centering
    \def\svgwidth{2.35in}
    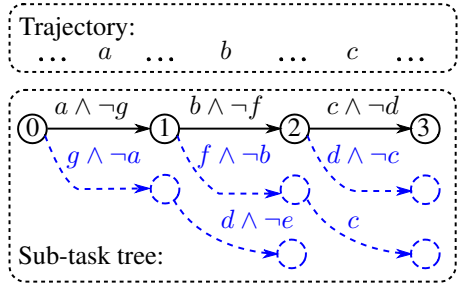
    \caption{\small Example of generating a sub-task tree as a new task based a collected trajectory. The blue part is composed by randomly generated sub-task sequences.}
    \label{fig:exp_relabel}
\end{figure}

In the level $k$ of the training curriculum, a collected trajectory $\zeta$ completes a sequence of sub-tasks $\tau:=\eta_0,\eta_1,\ldots,\eta_{l-1}$. A tree of randomly generated sub-tasks can be built based on $\tau$. Specifically, $\tau$ is first transformed into a tree with a single path, e.g., the black nodes and edges as shown in Figure \ref{fig:exp_relabel}. Second, for each sub-task $\eta_i (i<l)$, a sequence $\tilde{\tau}_i$ of sub-tasks with length $\le l-i\le k$ is randomly generated, where the length can be $0$, under the constraint that $\tilde{\tau}_i$ is never completed in trajectory $\zeta$. 
Then, the first sub-task in $\tilde{\tau}_i$, i.e., $\tilde{\tau}_i[0]$, and $\eta_i$ are modified to be {\it compatible} with each other, meaning that their positive propositions are different
and the positive propositions of $\eta_i$ ($\tilde{\tau}_i[0]$) are included in the negative propositions of $\tilde{\tau}_i[0]$ ($\eta_i$). For example, in Figure \ref{fig:exp_relabel} at node 1, the sub-task $b\wedge\neg f$ is compatible with $f\wedge\neg b$. Finally, this sequence $\tilde{\tau}_i$ is added to node $i$ as another branch. After repeating these steps for every $\eta_i$ in $\tau$, a sub-task tree can be generated and used a new task to relabel the trajectory $\zeta$, producing counterfactual experience to train the agent.
An example of generating a tree of sub-tasks as the new task is shown in Figure \ref{fig:exp_relabel}. 

\subsection{Avoiding Propositions}
\label{sec:avoid}
Every sub-task assigned by the high-level agent is a reach-avoid sub-task executed by the low-level policy, i.e., $\pi^l_{\omega}(\cdot|s,p_+,p_-,\phi')$, meaning that from a state $s$ the agent is required to avoid any propositions in $p_-$ before reaching propositions $p_+$ ($\bigwedge_{g_j\in p_+} g_j$ holds true), conditioned on completing the future task $\phi'$. In order to improve learning efficiency, the low-level policy in our method only performs avoidance in situations where there is a high likelihood of colliding with any propositions in $p_-$. The likelihood of colliding with a proposition $g$ in $p_-$ from current state $s$ can be assessed by the low-level value function $V^l_{\omega}(s,g,\varnothing,\varnothing)$ (denoted as $V^l_{\omega}(s,g)$ in the following), where $\varnothing$ refers to empty input. If $V^l_{\omega}(s,g)$ exceeds a threshold $\nu$, it indicates a high chance of colliding with proposition $g$. According to the design of training curriculum, for any $g\in\mathcal{P}_0$, $V^l_{\omega}(s,g)$ and the corresponding Q function $Q^l_{\omega}(s,g,a)$ are well trained in the first level. 

Specifically, define $k:=\arg\max_{g_k\in p_-} V^l_{\omega}(s,g_k)$. If its value $V^l_{\omega}(s,g_k)$ is below the threshold $\nu$ (e.g. $0.9$), it means that the agent is in the safe zone, and can take a goal-reaching action, i.e., $\pi^l_{\omega}(\cdot|s,p_+,p_-,\phi')$. Otherwise, the agent needs to select a safe action to avoid proposition $g_k$, i.e., $\arg\min_a Q^l_{\omega}(s,g_k,a)$, which moves the agent away from proposition $g_k$. 

With the guidance of safe actions, the frequency of violation of avoidance constraints (visiting propositions in $p_-$) can be reduced in every collected trajectory. Hence, since the low-level policy $\pi^l_{\omega}(\cdot|s,p_+,p_-,\phi')$ is trained by collected trajectories for solving reach-avoid task $(p_+,p_-)$ via PPO, the data-efficiency of low-level training can be improved.

\begin{figure}
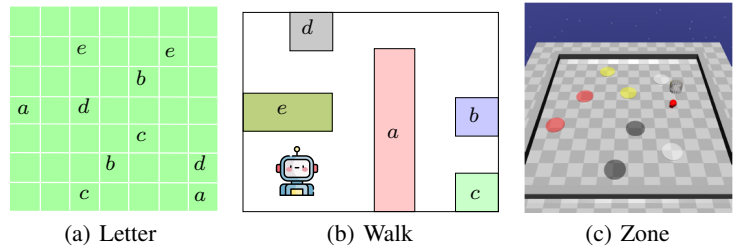

    \centering
    \subfigure[Letter]{
        \fontsize{8pt}{10pt}\selectfont
        \def\svgwidth{1.1in}
        \input{letter_map.pdf_tex}
        \label{fig:letter_map_app}
    }
    \subfigure[Walk]{
        \centering
        \fontsize{8pt}{10pt}\selectfont
        \def\svgscale{1.2}
        
        \label{fig:walk_env_app}
    }
    \subfigure[Zone]{
        \centering
        \includegraphics[width=1.1in, height=1.1in]{robot_zone.png}
        \label{fig:robot_zone_app}
    }
    \caption{Environments}
    \label{fig:envs_app}
\end{figure}

\section{Environments}
\label{sec:app_envs}
\noindent
{\bf Letter.} This environment is a $n \times n$ grid game with multiple letters positioned on the map. Among the $n^2$ cells, $m$ are assigned one of $k$ unique letters (where $m > k$). Certain letters may appear in multiple cells, allowing the agent to encounter other letters on its way to a target letter. Figure \ref{fig:letter_map_app} illustrates an example layout with $n=7$, $m=10$, and $k=5$. At each step, the agent can move in any cardinal direction (up, down, left, or right). The agent receives a task specification and is assumed to have an egocentric view of the entire grid and its letters through image-based observation.

\noindent
{\bf Walk.} 
This environment is a two-dimensional continuous world. As illustrated in Figure \ref{fig:walk_env_app}, the agent navigates through this world to visit various colored regions, with each color representing a different letter. When the agent enters a region, the corresponding letter is generated by a labeling function and observed by the agent. The agent’s state consists of the coordinates of its position, starting at $(0,0)$. The environment's dynamics are defined as $x' = x + a/10$, where $x \in [-1,10]^2$ represents the state vector, and $a \in [-1,1]^2$ denotes the action. The agent has no prior knowledge of the regions’ positions.

\noindent
{\bf Zone.} 
This is a robotic environment featuring continuous action and state spaces, adapted from OpenAI’s Safety Gym \citep{ray2019benchmarking}. As shown in Figure \ref{fig:robot_zone}, the environment is a 2D plane with 8 colored zones, where each color represents a proposition in the task specification, and there are two zones per color. We use a Safety Gym robot called Point, equipped with one actuator for turning and another for moving forward or backward. The agent can observe LiDAR information of the surrounding zones, using this indirect geographical data to visit or avoid specific zones to meet task requirements defined in SPECTRL. At the start of each episode, the positions of both the zones and the robot are randomized.

\begin{figure}
    \centering
    \subfigure[Letter]{
        \centering
        \includegraphics[width=1.35in]{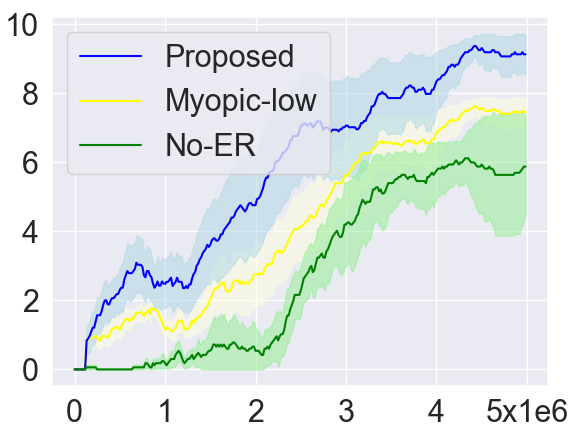}
        \label{fig:letter_abl}
    }
    \subfigure[Walk]{
        \centering
        \includegraphics[width=1.35in]{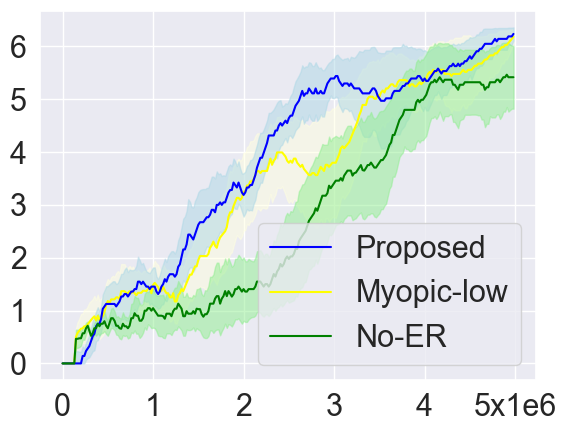}
        \label{fig:walk_abl}
    }
    \subfigure[Zone]{
        \centering
        \includegraphics[width=1.35in]{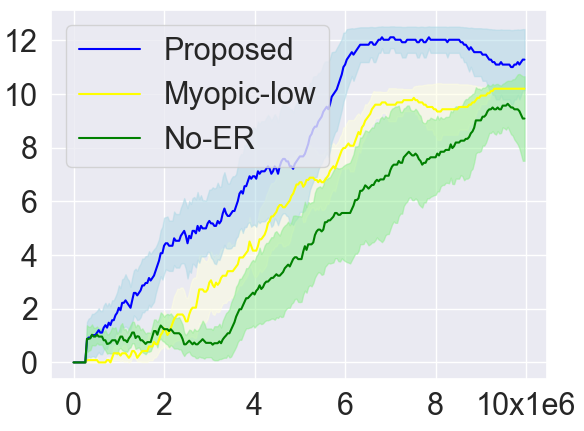}
        \label{fig:zone_abl}
    }
    \caption{Ablation study. The x-axis is the number of environmental samples used for training. The y-axis is the average return per episode.}
    \label{fig:ablation}
\end{figure}

\section{Ablation Study}
\label{sec:ablation}
In this section, we assess the effectiveness of key components within the proposed framework. First, we evaluate the baseline method "Myopic-low," which removes the embedding from the low-level module inputs. This setup tests the impact of accounting for future sub-task dependencies at the low level. As shown in Figure \ref{fig:ablation}, "Myopic-low" shows declines in both learning efficiency and average return, highlighting the importance of conditioning on the embedding of future sub-tasks at the low level. This is because this embedding enables low-level policies to consider dependencies on upcoming sub-tasks and preserves the symmetry between the low- and high-level modules.

Additionally, Baseline-1 from Section \ref{sec:results} replaces the high-level module with a conventional value iteration planning method, effectively evaluating the role of future dependencies at the high level. So, there is no need to have another baseline "Myopic-high" evaluated here.

Second, the baseline method "No-ER" excludes experience relabeling (ER) from the proposed framework, allowing us to evaluate the impact of ER on learning efficiency. As shown in Figure \ref{fig:ablation}, ER significantly enhances learning efficiency. This is because the different GNNs used in both the low-level and high-level modules require training across diverse task forms to effectively incorporate dependencies of future sub-tasks. Furthermore, Figure \ref{fig:ablation} indicates that ER contributes more to improving learning efficiency than conditioning on future dependencies at the low level.

\section{Training Details}
\label{sec:hyper}
\subsection{Task Generation}
\label{sec:task_generation}
The testing tasks are all randomly generated DAGs. On every edge of a generated DAG, the symbol to achieve is randomly selected from $\mathcal{P}$ and negative symbols for safety condition are also randomly selected from $\mathcal{P}$.

\subsection{Reward Function of Subgoals}
\label{sec:high-reward}
In addition to successful task completion, the agent may get extra rewards when some subgoals are achieved in certain order. This introduces sub-task dependencies in the subgoal level. The conditions for getting extra rewards are: in Letter domain, first go to "a" then "d" for $+5$ reward, or first go to "b" while avoiding "a" then "e" for $+10$ reward; in Walk domain, first "d" while avoiding "e" then "b" for $+5$ reward, or first "b" while avoiding "a" then "c" for $+10$ reward; in Zone domain, first visit "green" then "red" zone for $+10$ reward, or first visit "red" while avoiding "yellow" then "blue" zone for $+20$ reward.

\subsection{High-level Architecture}
\label{sec:hyper-high}
We use PPO algorithm to train the high-level module. We used the same policy network $\pi^h_{\theta}$ (3 fully-connected layers with $[128, 128, 128]$ units and ReLU activations) and critic network $V^h_{\theta}$ (3 fully-connected layers with $[128, 128, 1]$ units and Tanh activation) model for Letter, Walk and Zone domains. 

The binary encoding of a sub-task is the concatenation of binary encodings of $p_+$ and $p_-$. The encoding of $p_+$ ($p_-$) is a $|\mathcal{P}_0|$-dimensional binary vector, whose $k$-th element is $1$ whenever the $k$-th proposition in $\mathcal{P}_0$ is included in $p_+$ ($p_-$). Hence, the binary encoding of any sub-task has $2|\mathcal{P}_0|$ elements.

The output of the policy network is the binary encoding of a feasible subtask, which is discrete in every domain. Specifically, the policy’s output was passed through a logit layer before softmax. Since, the task DAG is given and known to the agent, the feasible choices of next sub-tasks to complete are straightforward to obtain. Hence, at the output distribution of the policy, infeasible sub-tasks are masked to make sure that only feasible ones can be sampled. 

The encoder $\mathcal{E}_{\theta}$ is determined by the observation space of the working domain: in Letter we used a 3-layer convolutional neural network with 16, 32 and 64 channels, kernel size of $2\times 2$ and stride of 1, and in Walk and Zone we used a 2-layer fully-connected network with $[128, 128]$ units and ReLU activations. In addition, the output is further mapped by a single linear layer to the latent representation vector, representing the input state $s_t$ in the latent space. The dimension of latent space is $64$ for Letter and Walk and $128$ for Walk and Zone. The latent transition model $\mathcal{T}_{\theta}$ is realized by a 3-layer fully connected network with $[128, 128, 128]$ units and ReLU activations. 

The GNN component used in Section \ref{sec:high-agent} to extract the embedding vector for future sub-tasks is a GNN architecture with $T=8$ message passing steps and $F$-dimensional node features, i.e., $\tilde{h}_k\in\mathbb{R}^{F}$. Here, one message passing step is one layer or iteration of applying functions $\mathcal{M}_{\theta}$ and $\mathcal{U}_{\theta}$ described in Section \ref{sec:high-agent}. Both functions $\mathcal{M}_{\theta}$ and $\mathcal{U}_{\theta}$ are realized by a 2-layer fully connected MLP with $F$ units and ReLU activations. Specifically, $F=64$ for Letter and Zone domains while $F=32$ for Walk domain. In order to reduce the number of trainable parameters, $\mathcal{M}_{\theta}$ and $\mathcal{U}_{\theta}$ are used and shared in every iteration of message passing. We observe that the performance can be improved by concatenating the embedding of a node at each iteration with its one-hot encoding of subgoal before passing it to the neighboring nodes for aggregation.

\subsection{Low-level Architecture}
\label{sec:hyper-low}
In the low-level module, the encoder for processing observations from the environment has the same architecture as the encoder used in the high level. 

In addition, we use the same actor network $\pi^l_{\omega}$ (3 fully-connected layers with $[64, 64, 64]$ units and ReLU activations) and critic network $V^l_{\omega}$ (3 fully-connected layers with $[64, 64, 1]$ units and Tanh activation) model for Letter, Walk and Zone domains. In Letter domain, the actor's output is a distribution of cardinal actions produced by a softmax layer, while in Walk and Zone domains, the actor produces a Gaussian action distribution and parametrizes its mean and standard deviation by sending the actor’s output to two separate linear layers.

For processing the DAG of the remaining task, we use a GCN architecture with $8$ message passing steps and $32$-dimensional node features. To reduce the number of trainable parameters, the weight matrix is shared across iterations for each edge type.

\end{document}